\documentclass[letterpaper, 10 pt, conference]{ieeeconf}  

\IEEEoverridecommandlockouts                              

\usepackage{amsmath} 
\usepackage{algorithmic}
\usepackage{amsfonts}
\usepackage{bbm}
\usepackage{bbold}
\usepackage{mathtools}
\usepackage{cite}
\usepackage[table]{xcolor}
\usepackage{graphicx}
\usepackage{array,booktabs,arydshln}
\usepackage{hyperref}
\usepackage{algorithm}
\usepackage{caption}

\definecolor{green}{RGB}{3,200,15}

\title{\LARGE \bf
Learning Real-world Autonomous Navigation by \\Self-Supervised Environment Synthesis 
}

\author{Zifan Xu$^{\ssymbol{1}}$, Anirudh Nair$^{\ssymbol{1}}$, Xuesu Xiao$^{\ssymbol{2}, \ssymbol{3}}$, and Peter Stone$^{\ssymbol{1},  \ssymbol{4}}$
\thanks{$^{\ssymbol{1}}$Department of Computer Science, The University of Texas at Austin $^{\ssymbol{2}}$Department of Computer Science, George Mason University $^{\ssymbol{3}}$Everyday Robots $^{\ssymbol{4}}$Sony AI \scriptsize\texttt{ \{zfxu, ani.nair\}@utexas.edu, xiao@gmu.edu, pstone@cs.utexas.edu} }
}

\makeatletter
\newcommand{\ssymbol}[1]{^{\@fnsymbol{#1}}}
\makeatother

\begin{document}

\maketitle
\thispagestyle{empty}
\pagestyle{empty}

\begin{abstract}
Machine learning approaches have recently enabled autonomous navigation for mobile robots in a data-driven manner. Since most existing learning-based navigation systems are trained with data generated in artificially created training environments, during real-world deployment at scale, it is inevitable that robots will encounter unseen scenarios, which are out of the training distribution and therefore lead to poor real-world performance. On the other hand, directly training in the real world is generally unsafe and inefficient. To address this issue, we introduce Self-supervised Environment Synthesis (\textsc{ses}), 
in which, after real-world deployment with safety and efficiency requirements, autonomous mobile robots can utilize experience from the real-world deployment, reconstruct navigation scenarios, and synthesize representative training environments in simulation. Training in these synthesized environments leads to improved future performance in the real world. 
The effectiveness of \textsc{ses} at synthesizing representative simulation environments and improving real-world navigation performance is evaluated via a large-scale deployment in a high-fidelity, realistic simulator\footnote{Due to the lack of access to large-scale real-world deployment data, we use simulated Matterport environments \cite{Matterport3D} as a surrogate of the real world.} and a small-scale deployment on a physical robot.

\end{abstract}

\section{INTRODUCTION}
\label{sec::introduction}

While classical navigation systems have been able to move mobile robots from one point to another in a collision-free manner for decades \cite{quinlan1993elastic, fox1997dynamic}, learning-based approaches to navigation have recently gained traction \cite{xiao2020motion} due to their ability to learn navigation behaviors purely from data without extensive engineering effort. For example, learned navigation systems can learn from human demonstrations \cite{pfeiffer2017perception} or self-supervised trial and error \cite{zhang2017deep}; they can learn navigation cost functions that consider social norms and human preferences \cite{perez2018teaching}. 
They can also be combined with classical navigation systems to assure navigation safety and enable adaptive behaviors in different scenarios \cite{ma2021navtuner, daftry2022mlnav, xu2021applr, xiao2021appl}. 

\begin{figure}[t]
\vspace{5pt}
  \centering
  \includegraphics[width=0.9 \columnwidth]{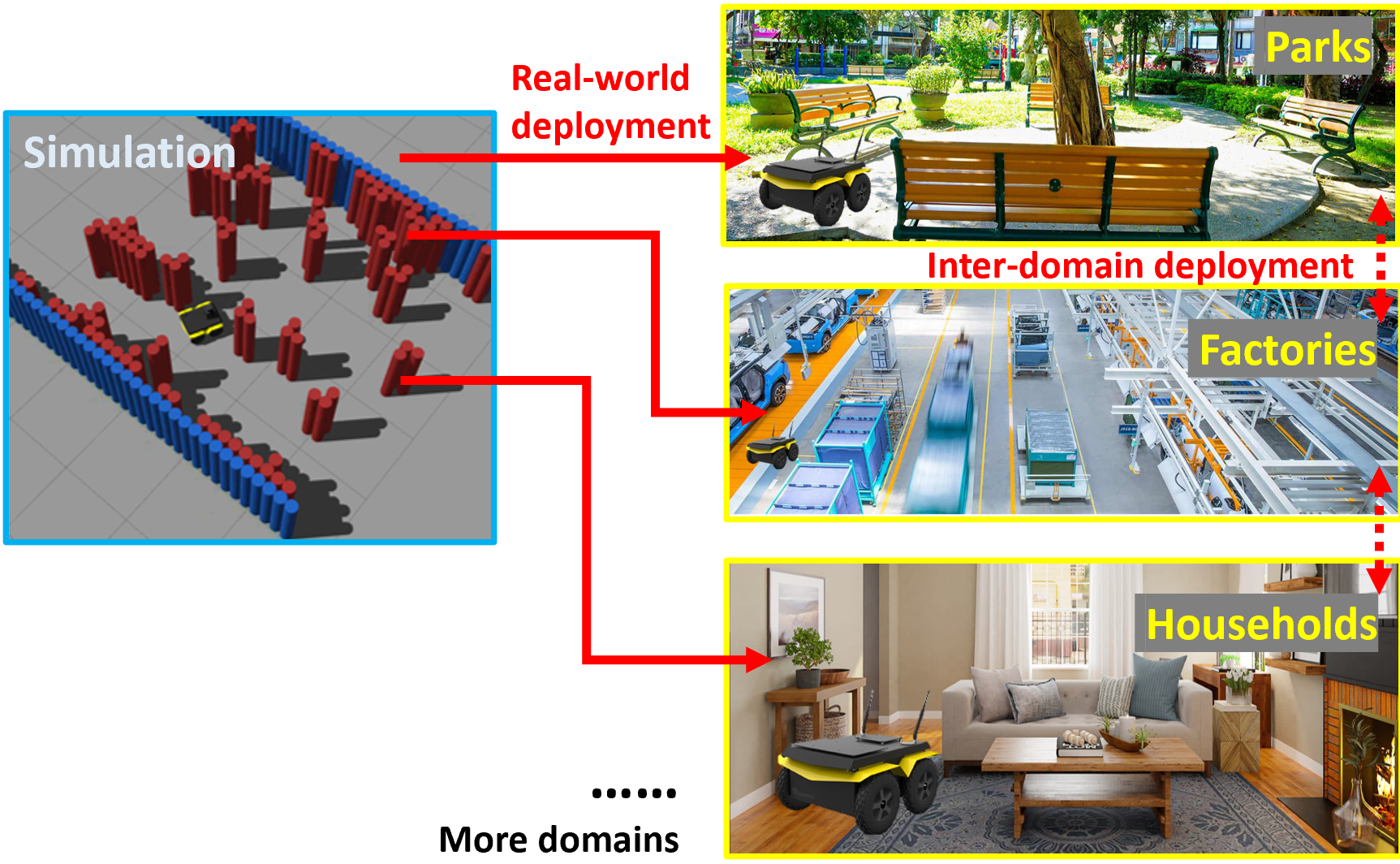}
  \caption{A navigation policy trained in simulation is expected to be deployed in completely different domains of navigation environments in the real world (e.g., households, factories, and parks). The policy may also face different real-world inter-domain deployments, in which a navigation policy learned in one real-world domain will be deployed in another.}
  \label{fig::deploy}
  \vspace{-15pt}
\end{figure}

Due to the expense of trial-and-error training in the real world (e.g., safety concerns and sample efficiency), most navigation behaviors are learned in artificially created environments in simulation, which may not generalize well to the real world (see Fig. \ref{fig::deploy}). Despite efforts to create simulation environments similar to the real world or enable efficient sim-to-real transfer, it is inevitable that robots will encounter unfamiliar scenarios, especially in large-scale real-world deployments.

The goal of this work is to improve real-world autonomous navigation with safety and efficiency requirements based on mobile robots' own navigation experiences during actual deployment. These conservative, potentially suboptimal, real-world experiences (without risky real-world exploration) may not be sufficient to directly train an RL agent, but may be sufficient to reconstruct the real-world navigation scenarios which an RL agent can interact with and actively explore in simulation. On the other hand, given the large amount of real-world deployment experiences available to many robots deployed in the field (consider 7 million connected iRobot Roombas vacuuming homes day to day), it is infeasible to reconstruct all these deployment environments and train in simulation on a daily basis.

With this motivation in mind, this paper introduces Self-supervised Environment Synthesis (\textsc{ses}), which enables mobile robots deployed in the field to first reconstruct navigation scenarios from experiences and then synthesize a representative set of simulation environments that is feasible for RL training. Training in these simulated environments enables robots to learn to address real-world challenges that they are likely to encounter.

Importantly, the distribution of real-world navigation scenarios is often unbalanced, including mostly trivial open scenarios. Therefore, we use an efficient strategy that filters out the trivial scenarios by a measure of navigation difficulty and focus learning on the challenging navigation scenarios. To synthesize the training environment set from the reconstructed challenging navigation scenarios, 
three different environment synthesis approaches—Generative Adversarial Networks (\textsc{gan}), K-means clustering with Principle Component Analysis (\textsc{pca}), and random sampling—are employed to represent the challenging scenarios with a concise training environment set that is feasible for an RL agent to learn from. We denote the pipelines with three environment synthesis approaches as \textsc{ses-gan}, \textsc{ses-pca} and \textsc{ses-rs} respectively.
We evaluate the three \textsc{ses} pipelines in Matterport, a dataset of many simulated realistic household environments (which serves as a surrogate of real world), and show that \textsc{ses} improves the deployment in these environments compared to policies trained in artificially generated environments \cite{perille2020benchmarking}, and the best pipeline \textsc{ses-gan} generates more representative training environments and enables better deployment in Matterport than the pipelines with other synthesis approaches. 

\section{RELATED WORK}


\subsection{Classical and Learning-Based Navigation}
Mobile robot navigation has been investigated by roboticists for decades \cite{quinlan1993elastic, fox1997dynamic}. Classical approaches can move robots from one point to another with a reasonable degree of confidence that they won't collide with any obstacles. However, these approaches require extensive engineering to develop in the first place and to adapt to different environments. Moreover, when encountering an environment in which a robot has failed or achieved suboptimal behavior before, without re-engineering the system, the robot will likely repeat the same mistake again. 

Inspired by the success of machine learning in other domains, roboticists have also applied machine learning to autonomous navigation \cite{xiao2020motion}. Most learning approaches to navigation adopt an end-to-end approach, i.e., learning a mapping from perceptual input directly to motion commands. Such approaches require comparatively less engineering effort, and learn navigation behaviors purely from data \cite{pomerleau1989alvinn}, e.g., from expert demonstrations \cite{pfeiffer2017perception, xiao2021learning} or from trial and error \cite{zhang2017deep, karnan2021voila}. However, these approaches often lack safety guarantees and explainability, as provided by their classical counterparts. Therefore, roboticists have also investigated more structured ways of integrating learning with classical navigation, such as learning local planners \cite{xiao2021toward, xiao2021agile, wang2021agile}, terrain-based cost functions \cite{wigness2018robot}, planner parameters \cite{xu2021applr, xiao2021appl}, driving styles \cite{kuderer2015learning}, or social norms \cite{perez2018teaching}. Their success notwithstanding, learning-based navigation approaches inherit one drawback from machine learning approaches in general: poor generalizability when facing out-of-distribution data. When deployed in the real world, especially at large scale, it is inevitable that mobile robots will encounter scenarios that are not included in their training distribution. 

\textsc{ses} combats classical navigation's inability to improve from experience \cite{liu2021lifelong} and learning approaches' poor generalizability to real-world scenarios. It improves navigation by synthesizing training environments from real-world deployment experiences.

\subsection{Sim-to-real Transfer} 
Limited by the safety and efficiency requirements in the real world, a learning-based navigation system is usually trained in simulation. However, policies trained in simulation can perform poorly in the real world due to the mismatch between the simulation and the real world. This phenomenon is commonly referred to as the \emph{sim-to-real gap}. 

One major source of the \emph{sim-to-real gap} is the discrepancies between the sensor input rendered in simulation and the real robot's sensors. For example, to bridge the gap between real-world and synthetic camera images of a robotic system, prior work has employed techniques such as pixel-level domain adaptation, which translates synthetic images to realistic ones at the pixel level \cite{bousmalis2018using, rao2020rl}. These adapted pseudo-realistic images bridge the \emph{sim-to-real gap} to some extent, so policies learned in simulation can be executed more successfully on real robots by adapting the images to be more realistic. Another source of the  \emph{sim-to-real gap} is caused by dynamics mismatch between simulation and the real world e.g., due to an imperfect physics engine. A common paradigm to reduce the dynamics mismatch is Grounded Simulation Learning (GSL), which either directly modifies (i.e., grounds) the simulator to better match the real world \cite{farchy2013humanoid}, or learns an action transformer that induces simulator transitions that more closely match the real world~\cite{IROS20-Karnan, MACHINELEARNING21-karnan}.

In contrast to the two \emph{sim-to-real gaps} introduced above, this work addresses a gap caused by the environmental mismatch (e.g., differences in the configurations and shapes of obstacles, and start-goal locations). \textsc{ses} can be thought of as an environmental grounding method that minimizes the differences in navigation environments between simulation and the real world based on the navigation experiences collected during real-world deployment.

\section{APPROACH}
\label{sec::approach}
In this section, We first formulate large-scale real-world navigation as a multi-task RL problem in an unknown navigation domain, which is defined as a distribution of navigation tasks. Sec. \ref{subsec::nav_task} formally defines the navigation task and describes how a distribution of navigation tasks forms a navigation domain. Then, Sec. \ref{sec::sshs} and \ref{sec::envsyn} discuss the two stages of \textsc{ses}: real-world navigation domain extraction from real-world deployment data and environment synthesis that generates a representative set of navigation tasks. The whole pipeline of \textsc{ses} is summarized in Alg. \ref{alg::ses}.

\subsection{Navigation Task and Navigation Domain}
\label{subsec::nav_task}
We focus on a standard goal-oriented navigation task, in which a robot navigates in a navigation environment $e$ from a provided starting pose $\alpha$ to a goal pose $\beta$. Each navigation task $T$ is instantiated as a tuple $T = (e, \alpha, \beta)$. In real-world applications, robots are not deployed to navigate in one single environment or with the same start and goal all the time. Instead, actual deployments in the real world usually entail a distribution over multiple environments with many start and goal poses. In this case, we represent the real-world deployment as a navigation domain $p_{\text{real}}$ defined as a distribution of navigation tasks $p_{\text{real}}(T)$. 
\textsc{ses} generates a new navigation domain $p_{\textsc{ses}}$ in simulation that, with a limited number of tasks, models the distribution of tasks in ${p_{\text{real}}}$ so that the navigation performance of policies trained in $p_{\textsc{ses}}$ is likely to be strong in the real-world navigation domain ${p_{\text{real}}}$.
\textsc{ses} uses \textsc{applr} \cite{xu2021applr} as the learning approach that solves the navigation tasks by training a parameter policy that dynamically adjusts the hyper-parameters of a classical navigation stack. Although our implementation of \textsc{ses} is based on a specific learning approach (\textsc{applr}), we leave the formulation sufficiently broad so that \textsc{applr} can be replaced with any RL-based approach to autonomous navigation.

\subsection{Real-world Navigation Domain Extraction}
\label{sec::sshs}
\textsc{ses} seeks to explore the distribution of the real-world navigation domain during real deployment of an existing navigation policy $\pi_0$ (e.g., a policy pretrained on artificially created environments or a classical navigation system). In each real-world deployment, a navigation task $T_n \sim p_{\text{real}}(T)$ is sampled from the real-world navigation domain. At each deployment time step $k$, \textsc{ses} records the sensory input $x_k$, the robot's position $s_k$ in the world frame, and the number of suboptimal or failed navigation behaviors $c_k$ (negative velocity command is used as the indication of failure in this paper).\footnote{The navigation policy used in this paper is based on a classical navigation system \textsc{dwa}, in which, suboptimal recovery behavior assigns negative linear velocity to the robot} Each deployment will return a trajectory $\tau = (x_0, s_0, c_0, ..., x_k, s_k, c_k,..., x_K, s_K, c_K)$ with $K$ being the total number of steps during a deployment. Then, the trajectories are segmented into different scenarios $\eta$ as follows: (1) starting from an \emph{initial} step $(x_i, s_i, c_i)$, iterate the trajectory until a \emph{final} step $(x_f, s_f, c_f)$ at which the robot is 5m away from the \emph{initial} step; (2) record a scenario $\eta$ as a sub-trajectory of all the steps between step $i$ and step $f$;  (3) after one scenario is recorded, set the \emph{final} step as the new \emph{initial} step for the next iteration. The procedure is repeated until the whole trajectory is processed. Assuming a total of $N$ rounds of deployments with trajectories segmented into $J$ scenarios, a trajectory set $\{\tau_n\}_{n=1}^N$ is collected and converted into a scenario set $\{\eta_j\}_{j=1}^J$. 

\begin{figure}
\vspace{5pt}
  \centering
  \includegraphics[width=0.7 \columnwidth]{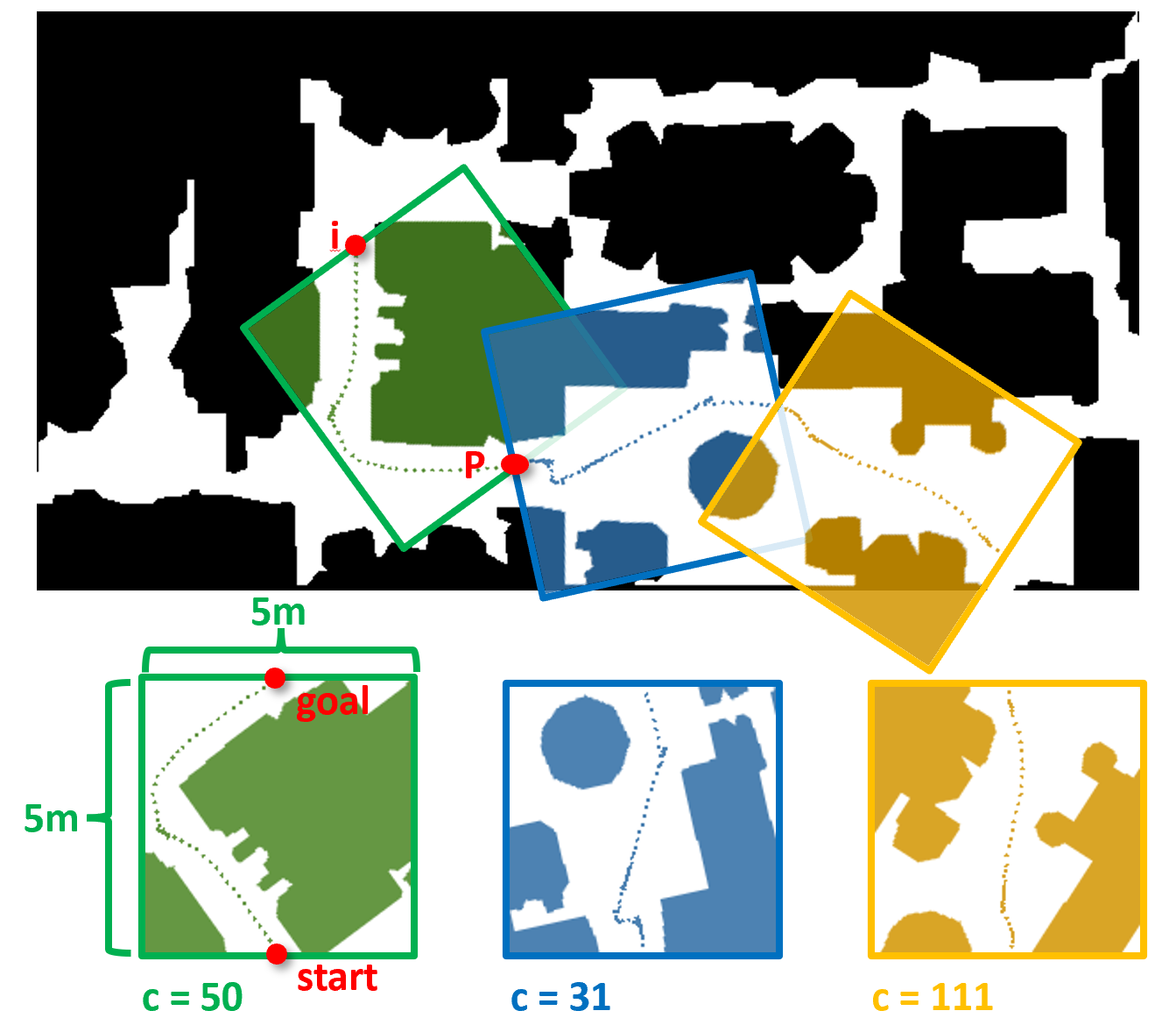}
  \caption{An example of trajectory segmentation and environment construction: dashed line denotes the actual trajectory of the robot. Three scenarios are segmented and constructed as navigation environments at the bottom. Red dots mark the positions of the start and goal in the environments. The numbers of suboptimal behaviors $c$ are noted at the bottom.} 
  \label{fig::real_gan}
  \vspace{-20pt}
\end{figure}

For each of the scenarios, we reconstruct a  5m-by-5m navigation environment with start and goal set to be the positions of the initial and final steps, and we re-orient the environment so that the start and goal are located at the middle points of the bottom and top edges respectively. The obstacles and free spaces can be reconstructed from the recorded sensory inputs or from available maps. The constructed environment set is denoted as $\{e_j, \bar{c}_j\}_{j=1}^{J}$. Each environment $e_j$ is associated with a total number of suboptimal navigation behaviors $\bar{c}_j$ between the \emph{initial} step and \emph{final} step. Fig. \ref{fig::real_gan} shows an example of such a trajectory segmentation and environment construction process. Each environment is represented as a 30$\times$30 binary matrix with each element denoting the occupancy state of a 1.5m$\times$1.5m space. The benefits of such scenario segmentation are: (1) encoding the start and goal into the orientation of the environment so that a single environment distribution can represent the real-world navigation domain distribution; (2) the segmented navigation environments have roughly the same length, which makes it easier for an RL agent to learn under roughly the same magnitudes of return and episode lengths across different environments.
Alg. \ref{alg::TDR} summarizes the above process of generating an environment set that represents the real-world navigation domain distribution.

A "raw" navigation domain $p_{\text{raw}}= U(\{e_i\}_{j=1}^J)$ can be defined as a uniform distribution over all these constructed navigation environments. However, it is likely that $p_{\text{raw}}$ includes many trivial navigation tasks (e.g., open spaces), whereas, efficient training should focus on the challenging navigation tasks. We use a threshold of 50 suboptimal behaviors within a scenario $\eta$ as an indication of easy and challenging environments. The threshold of 50 is empirically picked such that the trivial scenarios with mostly empty spaces will be filtered out while not losing the challenging scenarios.
Those challenging environments, $\{e_j, \bar{c}_j | \bar{c}_j > 50\}_{j=1}^J$, form a set for further environment synthesis. 
We denote the navigation domain with only challenging navigation environments as $p_{\text{chal}}=U(\{e_j | \bar{c}_j > 50\}_{j=1}^J)$.
Our empirical results indicate that training only on the navigation tasks with challenging scenarios instead of all the real-world scenarios is essential for improving real-world deployments.


\subsection{Environment Synthesis}
\label{sec::envsyn}
While simply using all the challenging environments $\{e_j | \bar{c}_j > 50\}_{j=1}^J$ can precisely represent $p_{\text{chal}}$ in real world, it is impractical to train an RL agent in all these environments. In this case, a smaller set of training environments that represents the real-world navigation domain distribution is required. Therefore, we propose three methods of environment synthesis that perform this training data selection process and generate a concise and representative training environment set. The three environment synthesis methods are: (1) Generative Adversarial Nets (\textsc{gan}) \cite{goodfellow2014generative}; (2) K-means clustering with Principal Component Analysis (\textsc{pca}) \cite{pedregosa2011scikit}; and (3) random sampling. 
We describe \textsc{gan}, K-means clustering with \textsc{pca}, and random sampling as follows.

\textbf{\textsc{gan}} learns a generator's distribution $p_g$ over all the possible environments $e \in \{1,0\}^{30\times30}$ to match a uniform distribution $p_{\text{chal}}$ over the challenging environment set. Given a prior on the input noise variables $p_z(z)$, a generator $G(z;\theta_g)$ is defined as a mapping from the input variable space to the environment space with $\theta_g$ representing the parameters of the network. A discriminator network $D(e;\theta_d)$ outputs the probability that $e$ comes from $p_{\text{chal}}$ rather than $p_g$. $D$ is trained to maximize the probability of assigning the correct label to both challenging environments and generated environments from $G$. Simultaneously, $G$ is trained to minimize $\log(1-D(G(z)))$. To summarize, $D$ and $G$ play the following two-player minimax game with value function $V(G,D)$: 
\begin{equation}
\begin{aligned}
\min_G \max_D V(D,G) = \mathbb{E}_{e \sim p_{\text{chal}}(e)}[\log(D(e))] + \\ \mathbb{E}_{z \sim p_z(z)}[\log(1-D(G(z)))].
\end{aligned}
\label{eqn::minimax}
\end{equation}
After a generator $G$ is learned, we query the generator $M$ times to draw $M$ environments as the training set, where $M$ is the size of the training environment set dependent on the training budget of the RL algorithm.

\textbf{K-means clustering with \textsc{pca}} first reduces the dimensionality of the original environment representation (30$\times$30 binary matrix) by \textsc{pca} so that each environment can be represented by their principle components (we empirically pick 100 principle components in our experiments because they can reasonably reconstruct the original environments). Then, K-means is employed to cluster the environments in the reduced space into $m$ clusters and sample $n$ environments from each cluster to select $M = m \times n$ representative environments in the challenging environment set. We use the principle components to reconstruct the representative environments and use them as the training set.

Random sampling simply uses $p_{\text{SES}} = p_{\text{chal}}$ the uniform distribution over all the challenging environments. Then, $M$ environments are drawn from $p_{\text{SES}}$ to form the training set.

The synthesized navigation domain is denoted as $p_{\text{SES}}$. Finally, an \textsc{applr} policy $\pi$ is trained in $p_{\text{SES}}$. In practice, the policy $\pi$ is initialized to be the deployed policy $\pi_0$ that was trained on the artificially created environments to speed up training.

\begin{algorithm}
	\caption{Self-supervised Environment Synthesis}
	\label{alg::ses}
	\begin{algorithmic}[1]
	    \REQUIRE{Original policy $\pi_0$, real-world navigation domain distribution $p_{\text{real}}$}, number of training environments $M$
	    \STATE{Raw navigation domain $p_{\text{raw}}$ $\leftarrow$ Real-world Navigation Domain Extraction based on $\pi_{0}$ and $p_{\text{real}}$} (Sec. \ref{sec::sshs})
	    \STATE Synthesized navigation domain $p_{\text{SES}}$ $\leftarrow $ Environment Synthesis based on  $p_{\text{raw}}$ (Sec. \ref{sec::envsyn})
	    \STATE $\{e_i\}_{i=1}^M$ $\leftarrow $ Sample $M$ environments from $p_{\text{SES}}$
	    \STATE $\pi \leftarrow$ Initialize with $\pi_{\text{0}}$, then train on $\{e_i\}_{i=1}^M$
	    \RETURN $\pi$
	\end{algorithmic}
\end{algorithm}

\begin{algorithm}
	\caption{Real-world Navigation Domain Extraction} 
	\label{alg::TDR}
	\begin{algorithmic}[1]
	    \REQUIRE{Original policy $\pi_0$, real-world navigation domain distribution $p_{\text{real}}$ and total number of deployments $N$.}
	    \STATE{$E = \emptyset$}
	    \FOR{$n \in \{1, ..., N$\}}
	        \STATE{$T_n \sim p_{\text{real}}(T)$}
	        \STATE{Trajectory $\tau_n \leftarrow$ deploy $\pi_0$ on $T_n$}
	        \STATE{Initialize step $i=0$ and the scenario set $\eta = \emptyset$}
	        \FOR{$x_k, s_k, c_k \in \tau_n$}
	            \STATE{$\eta \leftarrow \eta \cup \{(x_k, s_k, c_k)\}$}
	            \IF{Euclidean($s_k$, $s_i$) $> 5$}
	                \STATE{$e$ $\leftarrow$ Construct Environment from $\eta$}
	                \STATE{$\bar{c} = \sum_i^k c_i$}
	                \STATE{$E \leftarrow E \cup \{e, \bar{c}\}$}
	                \STATE{$i=k$,     $\eta =       \emptyset$}
	            \ENDIF
	        \ENDFOR
	    \ENDFOR
	    \RETURN{$E$}
	\end{algorithmic}
\end{algorithm}
\section{RESULTS}
\label{sec::results}
In this section, we describe an implementation of \textsc{ses} on a ground robot and show that \textsc{ses} can efficiently synthesize representative environments based on challenging navigation scenarios during real-world deployment and successfully improve navigation by learning from the synthesized environments. Due to the difficulty in accessing large-scale physical robot deployment data in the real world, we use Matterport \cite{Matterport3D}, a set of high-fidelity simulation environments as a surrogate for a large-scale real-world deployment (see Fig. \ref{fig:matterport} (d)). We assume that this environment set is not available before actual deployment, and the robot needs to learn through its actual deployment experiences in those environment to improve navigation. 

\subsection{Deployment in Matterport Navigation Domain}
\label{sec::sim}
We employ \textsc{ses} on a simulated ClearPath Jackal differential-drive ground robot. The specifications of the robot are kept the same as in \textsc{applr} \cite{xu2021applr}, including the \textsc{ros} \texttt{move\_base} navigation stack, the underlying classical local planner \textsc{dwa}, and the learned planner parameters.

\begin{figure}
\vspace{5pt}
    \centering
    \includegraphics[width=0.8 \columnwidth]{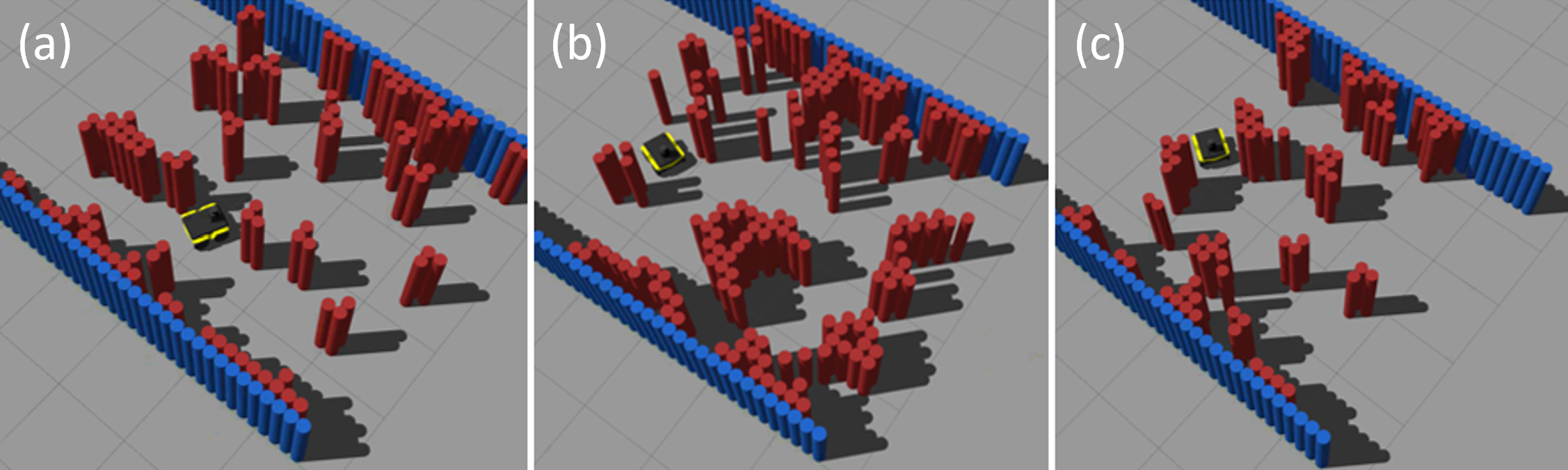}
    \centering
    \includegraphics[width=0.8 \columnwidth]{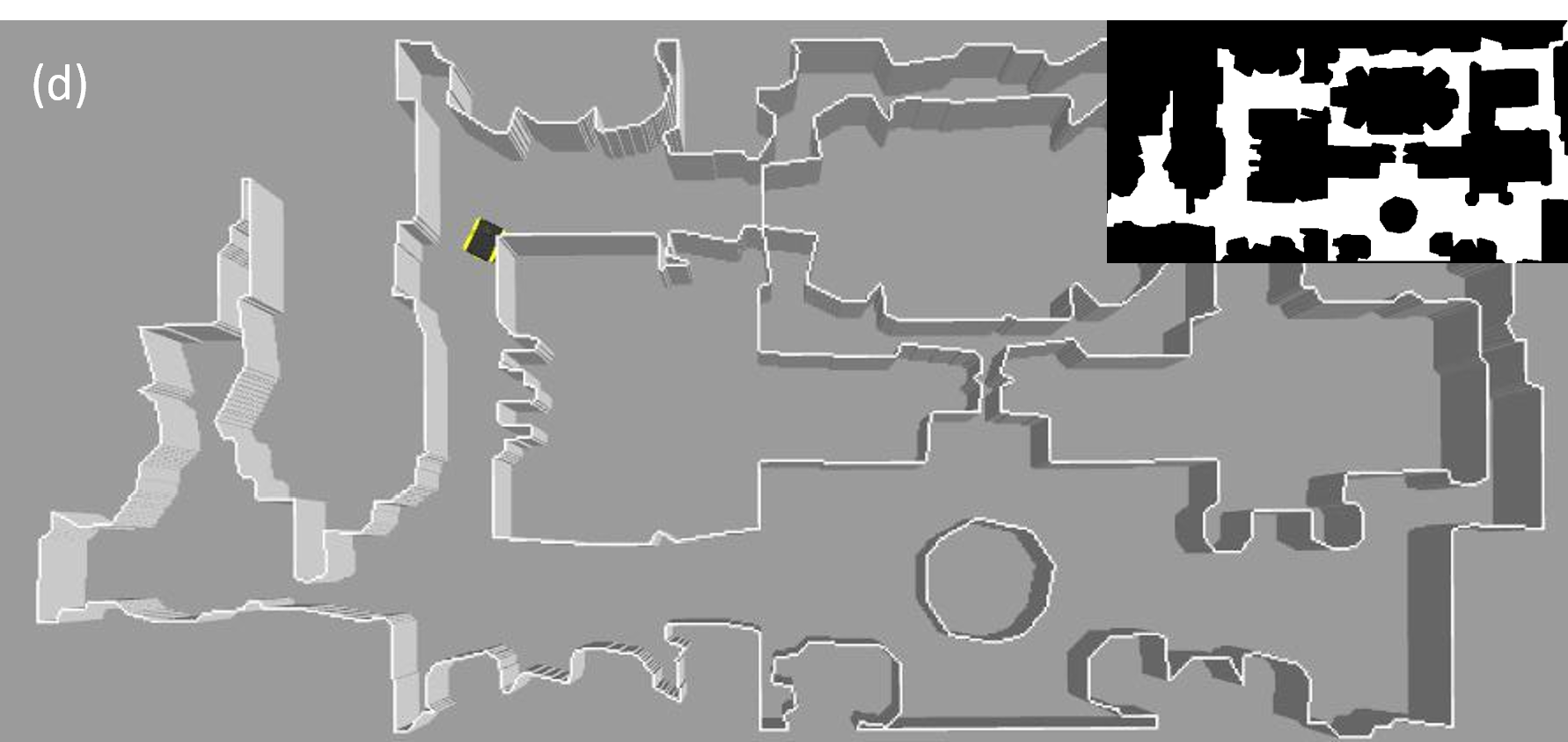}
    \caption{Examples of training and deployment environments. (a), (b), (c): example environments from the \textsc{barn} dataset; (d): a top-down view of a Matterport environment created from the occupancy map shown at the top-right corner. }
    \label{fig:matterport}
    \vspace{-15pt}
\end{figure}

The original \textsc{applr} policy was trained before any deployment in the \textsc{barn} dataset \cite{perille2020benchmarking}.  The dataset contains 300 simulated navigation environments randomly generated by Cellular Automata. Even though those environments are designed to be diverse enough to cover different difficulty levels from relatively open spaces to highly-constrained ones, when a controller learned on these environments is deployed in Matterport, the randomly generated environments in \textsc{barn} are found not to cover some specific obstacle shapes and arrangements encountered by the robot, and therefore lead to the failures at the deployment time. In Fig. \ref{fig:matterport}, (a)-(c) show three example \textsc{barn} environments, and (d) shows a Matterport environment where Jackal will be deployed. This realistic Matterport environment may contain challenging navigaiton scenarios that are unlikely to be generated by random sampling.

After training the original \textsc{applr} policy with the \textsc{barn} dataset, the robot will be physically deployed in the real world at scale. Due to the lack of access to large-scale real-world physical deployment data, we use 55 simulated house layouts from the Matterport dataset \cite{Matterport3D} as a surrogate for physical real-world deployment.
Fig. \ref{fig:matterport} (d) shows an example of such a deployment environment. We randomly sample 50 environments from the Matterport dataset \cite{Matterport3D} as the environments where Jackal will be actually deployed. During actual deployment, these environments serve as the Matterport navigation domain and navigation experiences in those environments during deployment provide data for \textsc{ses} to synthesize representative environments of the Matterport navigation domain and to improve navigation. We leave five environments to test the learned policy on these held-out Matterport environments.

\subsection{\textsc{ses} Implementation}
\label{sec::sesi}

\begin{figure}
  \centering
  \includegraphics[width=0.8 \linewidth]{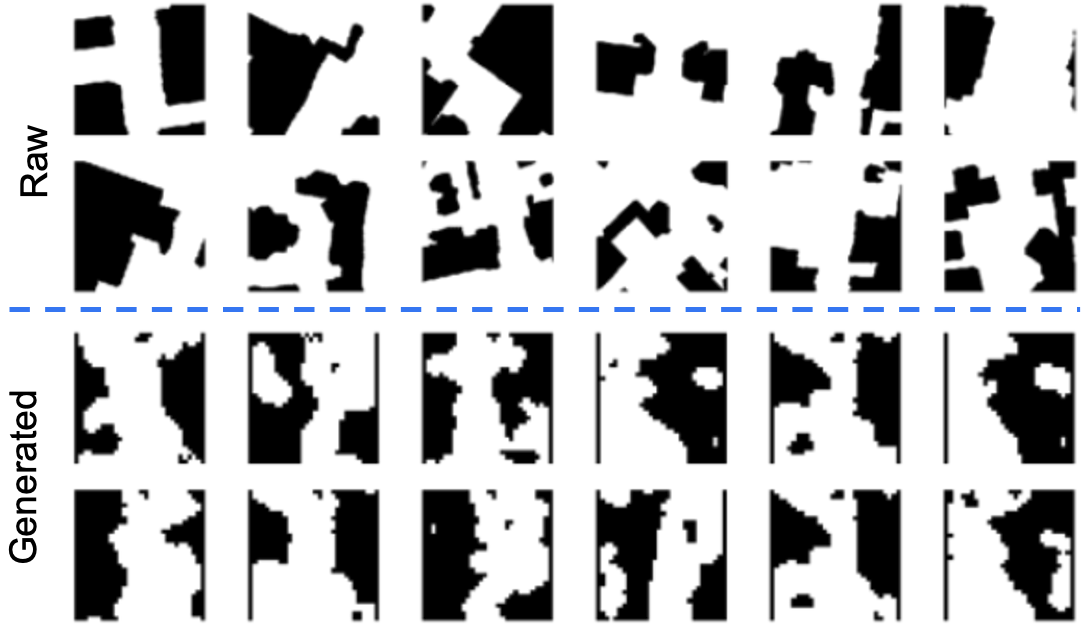}
  \caption{Examples of raw challenging environments (top) and generated environments (bottom).}
  \label{fig::real_gan_1}
  \vspace{-20pt}
\end{figure}

\textbf{Matterport navigation domain representation:} in the large-scale deployment in Matterport environments, the robot with the original policy $\pi_0$ trained in \textsc{barn} (artificially created navigation domain) is deployed 100 times to navigate between the starting and goal locations randomly sampled from all the navigable start-goal pairs in each Matterport environment, which accounts for 5000 deployments in total. We construct 16528 navigation scenarios from the segmented trajectories using the method described in Sec. \ref{sec::sshs} as $p_{\text{raw}}$ the distribution of the Matterport navigation domain. Then 3769 challenging environments are selected based on a threshold 50 of the negative linear velocity count, which forms $p_{\text{chal}}$. Fig. \ref{fig::real_gan_1} (left) demonstrates some examples out of the 3769 challenging environments. Each of the environments is represented as a binary matrix of size 30 $\times$ 30. Note that training in $p_{\text{chal}}$ with all 3769 environments is not computationally practical due the limit of CPUs that run the environments in parallel and the budget of total number of training steps. In the meantime, realistic large-scale robot deployments in the real world (in contrast to our surrogate) may generate orders of magnitude more challenging scenarios. 

\textbf{Environment synthesis:} given the extracted Matterport navigation domain $p_{\text{chal}}$, the environment synthesis method described in Sec. \ref{sec::envsyn} generates 100 training environments (the maximum number of environments that can be trained in parallel in our training pipeline) by using the challenging environments as an image dataset to train a \textsc{gan}. More specifically, we employ a normal distribution prior on the input noise variables $p_z(z):=\mathcal{N}(0,1)$, where $\mathcal{N}$ is the standard normal distribution with mean and standard deviation equal to 0 and 1 respectively. The generator $G(z;\theta_g)$ is represented by a multilayer perceptron (MLP) of 256, 512, and 1024-hidden-unit layers with batch normalization and leaky ReLU activation function. The generator outputs a 900-dimensional vector with a fully connected layer followed by a Tanh activation function. The output vector is further rendered into a $30 \times 30$ binary matrix using a threshold of 0. The discriminator $D(x;\theta_d)$ is represented by a second MLP of 512, 256, and 256-hidden-unit layers with leaky ReLU activation function. Each hidden layer is followed by a batch normalization layer and a dropout layer with a dropout rate equal to 0.5. The dropout serves as an ensemble strategy that prevents mode collapse in the generated images \cite{mordido2018dropout}. Fig. \ref{fig::real_gan} (right) shows samples of generated images drawn from the generator after training. The generator is queried 100 times to generate synthesized navigation domain $p_{\text{SES}}$ with 100 training environments, which is in the same order of magnitude as the \textsc{barn} dataset. We leave the investigation of how \textsc{ses} performs with different numbers and sizes of training environments as future work. 
To train the new policy $\pi$ that solves the navigation in challenging Matterport environments, we use the same set of hyper-parameters as when training $\pi_{0}$ except for a smaller training step of one fourth of the original 4M training step, because the pre-trained policy $\pi$ needs fewer training step to converge.

\begin{table}[bth!]
\vspace{5pt}
\centering
\caption{Average time costs and success rates}
\begin{tabular}{ccccc}
\toprule
  & Time cost (s) & Success rate (\%) &  \\ \midrule
\textsc{ses-gan} & 27.23 & 85.0 &  \\ 
\textsc{ses-pca} & 32.21 & 80.6 & \\
\textsc{ses-rs} & 35.92 & 71.0 & \\
\emph{Learn from scratch} & 30.42 & 74.3 & \\
Original policy & 36.33 & 71.9 &\\
\emph{\textsc{dwa} slow} & 43.76 & 75.5 &\\
\emph{\textsc{dwa} fast} & 31.72 & 67.2 &\\
\bottomrule
\end{tabular}
\label{tab::result}
\vspace{-15pt}
\end{table}

\subsection{Test Results}
\label{sec::test}
We train the new \textsc{applr} policy $\pi$ in $p_{\text{SES}}$ as created by the three \textsc{ses} methods: \textsc{ses-gan}, \textsc{ses-pca}, and \textsc{ses-rs}. We compare the methods with four baselines: the \textsc{applr} policy learned from scratch on the training environments generated by \textsc{gan}, the original \textsc{applr} policy $\pi_0$, and vanilla \textsc{dwa} with two sets of static hyper-parameters, one having default parameters and the other having a two times higher maximum linear velocity and sampling rate. We call the two classical motion planners \emph{\textsc{dwa} slow} and \emph{\textsc{dwa} fast} respectively. We use \emph{learn from scratch} to denote the first baseline which is designed to verify the importance of initializing the policy with the original policy $\pi_{\text{0}}$. Then we deployed the three adapted RL policies and four baselines in 5 held-out Matterport environments to test deployment performance. For each of the environments, we randomly sample five navigable start-goal pairs that have at least 5-meter start-goal distances 
, and deploy each policy 20 times for each pair. During deployment, we measure the success rate and the total time cost of the successful trials.

\begin{figure}
  \centering
  \includegraphics[width=0.9\columnwidth]{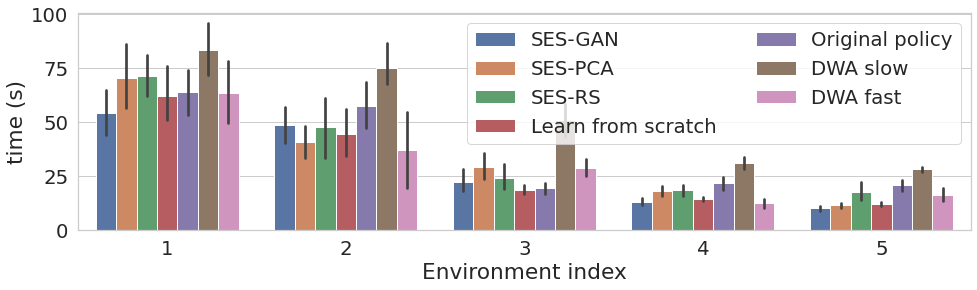}
  \centering
  \includegraphics[width=0.9\columnwidth]{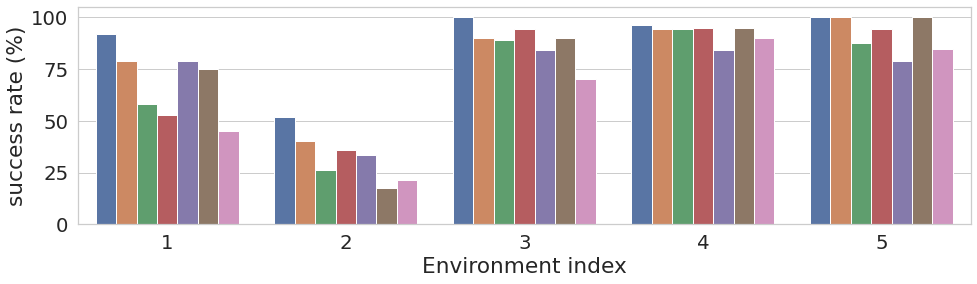}
  \caption{The average time cost (top) and the success rate (bottom) of 5 policies tested in 5 held-out environments.}
  \label{fig::result}
  \vspace{-5pt}
\end{figure}

Table \ref{tab::result} presents the average time cost and success rate over all trials.
\footnote{Standard deviation is not included in the table due to the large variance between different environments and start-goal pairs, but we analyze the variance of the performance in every individual environment in Fig. \ref{fig::result}.}
The policy trained by \textsc{ses-gan} achieves the best performances of a 27.23s average time cost and a 85\% success rate. \textsc{ses-pca} follows with a 32.21s average time cost and a 80.6\% success rate. \emph{Learn from scratch} shows worse performance on both metrics compared to \textsc{ses-gan} which indicates the importance of initializing the model with the pre-trained policy $\pi_{\text{0}}$. The policy trained by \textsc{ses-rs} performs comparably to the original policy that is not trained on the Matterport environments at all, which indicates the poor representation of the challenging Matterport environments with random sampling. While \emph{\textsc{dwa} slow} achieves an acceptable success rate of 75.5\%, it takes much longer (43.76s) to finish the navigation. On the other hand, \emph{\textsc{dwa} fast} can navigate relatively quickly, but more easily fails with the lowest success rate of 67.2\%. 

To analyze performance variance, we randomly select one start-goal pair from each environment and show the average time cost and success rate over 20 independent trials in Fig. \ref{fig::result}. The policies trained by \textsc{ses-gan} and \textsc{ses-pca}, in general, complete the navigation task more quickly and achieve higher success rate. Among five test environments, the \textsc{ses-gan} policy achieves the lowest time cost in three environments and the highest success rate in all five. \textsc{ses} methods tend to show larger improvements in the more difficult environments (1 and 2), but also larger performance variance.

\subsection{Physical Experiment}
\begin{figure}
  \centering
  \includegraphics[width=0.6\columnwidth]{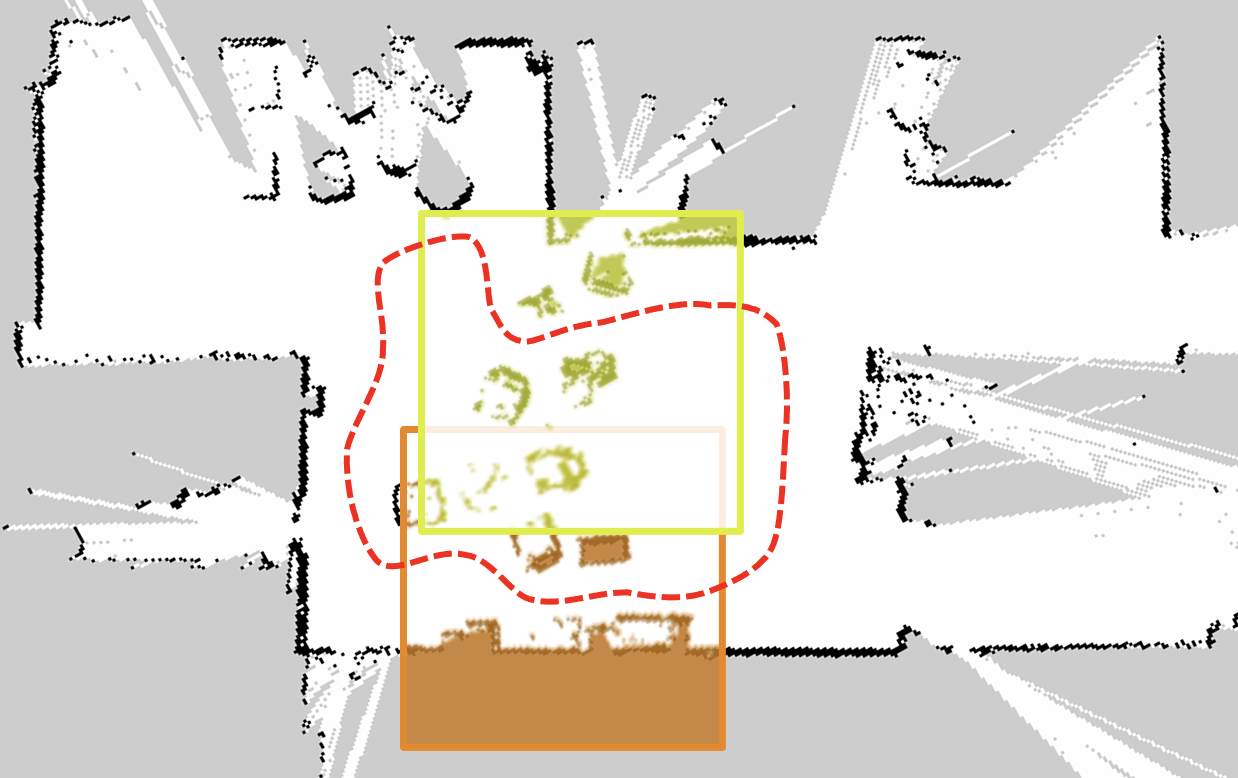}
  \caption{Real-world navigation path (red dashed line) and map. The orange and yellow areas mark the two challenging scenarios identified during the real-world deployment.}
  \label{fig::real_world_map}
  \vspace{-15pt}
\end{figure}

To verify that \textsc{ses} is potentially applicable to real-world robots, we deploy the pre-trained \textsc{applr} policy on a physical Jackal robot. The pre-trained policy is deployed in a real-world environment and a map of the environment is built during deployment using \texttt{gmapping} \cite{gmapping} (shown in Fig. \ref{fig::real_world_map}). Using the same criteria for suboptimal behavior as the Matterport experiments, the robot identifies two challenging scenarios shown by the orange and yellow areas in the map.

Considering the limited scale of our real-world deployment, we directly construct two synthetic environments which are identical to the challenging real-world scenarios (instead of using a \textsc{gan}) and fine-tune the pre-trained policy in them. We denote the policies before and after fine-tuning as \emph{original policy} and \textsc{ses} respectively. Table \ref{tab::real_world_result} reports the time cost averaged over 5 trials of both policies to traverse the two difficult real-world scenarios, where \textsc{ses} achieves significant improvement compared to the \emph{original policy}.

\begin{table}[H]
\centering
\caption{Average time cost of the tested policies in real-world challenging scenarios}
\begin{tabular}{ccccc}
\toprule
  & Scenario 1 & Scenario 2 \\ \midrule
Original policy & $18.5 \pm 7.6$ (s) & $38.4 \pm 9.8$ (s) \\ 
\textsc{ses} & $11.3 \pm 0.9$ (s) & $15.6 \pm 1.0$ (s) \\
\bottomrule
\end{tabular}
\label{tab::real_world_result}
\vspace{-10pt}
\end{table}
\section{CONCLUSION}
\label{sec::conclusion}
This paper introduces a self-supervised environment synthesis approach that improves the real-world navigation performance of a learning-based navigation system by synthesizing realistic simulation environments based on navigation experiences during real-world deployment. 
To address the inevitable environmental mismatch between simulation and real world deployment environments, especially for large-scale real-world deployment, \textsc{ses} can successfully identify challenging navigation scenarios during the deployment in high-fidelity Matterport environments and synthesize representative environments to be added to the training distribution. \textsc{SES} has also been demonstrated in a small-scale real-world deployment.
\newpage
\bibliographystyle{IEEEtran}
\bibliography{IEEEabrv,references}

\end{document}